\documentclass[a4paper,fleqn]{cas-dc}

\usepackage[numbers]{natbib}
\usepackage{framed,multirow}
\usepackage{lscape}
\usepackage{subcaption}
\usepackage{longtable}
\usepackage{academicons}
\usepackage{threeparttable}
\usepackage{placeins}
\usepackage{float}
\usepackage{graphicx}

\usepackage{url}
\usepackage{xcolor}

\def\tsc#1{\csdef{#1}{\textsc{\lowercase{#1}}\xspace}}
\tsc{WGM}
\tsc{QE}
\tsc{EP}
\tsc{PMS}
\tsc{BEC}
\tsc{DE}

\begin{document}
\let\WriteBookmarks\relax
\def\floatpagepagefraction{1}
\def\textpagefraction{.001}

\shorttitle{Reliable or Deceptive? Investigating Gated Features for Smooth Visual Explanations in CNNs}

\shortauthors{ P. Singh et~al.}

\title [mode = title]{Reliable or Deceptive? Investigating Gated Features for Smooth Visual Explanations in CNNs}

\author[1]{Soham Mitra}[type=editor,
                        auid=000,bioid=1,
                        prefix=,
                        orcid=,]
                        
\ead{cse21084@iiitkalyani.ac.in}
\affiliation[1]{organization={Indian Institute of Information Technology Kalyani},
    country={India}}

\author[2]{Atri Sukul}[type=editor,
                        auid=000,bioid=1,
                        prefix=,
                        orcid=,]

\ead{as2511@cse.jgec.ac.in}
\affiliation[2]{organization={Jalpaiguri Government Engineering College},
    country={India}}

\author[3]{Swalpa Kumar Roy}[type=editor,
                        auid=000,bioid=1,
                        prefix=,
                        orcid=,]

\ead{swalpa@agemc.ac.in}
\affiliation[3]{organization={Alipurduar Government Engineering and Management College},
    country={India}}

\author[4]{Pravendra Singh}[type=editor,
                        auid=000,bioid=1,
                        prefix=,
                        orcid=0000-0003-1001-2219]

\cormark[1]

\ead{pravendra.singh@cs.iitr.ac.in}
\affiliation[4]{organization={Department of Computer Science and Engineering, Indian Institute of Technology Roorkee},
    city={Roorkee},
    state={Uttarakhand},
    country={India}}

\author[5]{Vinay Verma}[type=editor,
                        auid=000,bioid=1,
                        prefix=,
                        orcid=,]

\ead{vverma.vinayy@gmail.com}
\affiliation[5]{organization={Amazon},
    country={India}}

\cortext[cor1]{Corresponding author: Pravendra Singh}

\begin{abstract}
Deep learning models have achieved remarkable success across diverse domains. However, the intricate nature of these models often impedes a clear understanding of their decision-making processes. This is where Explainable AI (XAI) becomes indispensable, offering intuitive explanations for model decisions. In this work, we propose a simple yet highly effective approach, ScoreCAM++, which introduces modifications to enhance the promising ScoreCAM method for visual explainability. Our proposed approach involves altering the normalization function within the activation layer utilized in ScoreCAM, resulting in significantly improved results compared to previous efforts. Additionally, we apply an activation function to the upsampled activation layers to enhance interpretability. This improvement is achieved by selectively gating lower-priority values within the activation layer. Through extensive experiments and qualitative comparisons, we demonstrate that ScoreCAM++ consistently achieves notably superior performance and fairness in interpreting the decision-making process compared to both ScoreCAM and previous methods. Our code is available at \href{https://github.com/AquillaNatus/ScoreCAMpp}{this https URL}.
\end{abstract}


\maketitle

\section{Introduction}
In various computer vision tasks like image classification \cite{liu2021swin,simonyan2014very,dosovitskiy2020image}, object recognition \cite{dai2021dynamic,zhu2020deformable}, and semantic segmentation \cite{kirillov2023segment}, as well as tasks like image captioning \cite{li2022blip,li2022grounded} and visual question answering \cite{li2022blip,patro2020robust}, the convolutional neural networks (CNNs) and other deep networks have enabled remarkable progress. As a result, when modern intelligent systems experience malfunctions, they often fail dramatically, producing unexpected and confusing results without prior warning or explanation. This leaves users with outcomes that are difficult to understand and can cause significant confusion.

Numerous methods \cite{zhou2016learning,selvaraju2017grad,chattopadhay2018grad,fu2020axiom,wang2020score,jiang2021layercam,electronics12234846,10295912,10288391,10304247} have been proposed to explain how deep networks make decisions, aiming to address the vital need for interpretability. Developing methods that create saliency maps, revealing important areas or features in inputs contributing to network outputs, is a well-explored research area. These maps provide explanations that offer insights into the network's internal processes and reasoning. However, despite significant progress, current explanatory methods often struggle to capture the complex connections and interactions within deep networks, limiting their interpretability. While they can identify critical areas, they often fall short in fully explaining how these areas influence the decision-making process. This constraint reduces the effectiveness of these systems and erodes user trust because they do not provide sufficient detail or contextual understanding

\begin{figure*}[t]

\centering

\includegraphics[scale=0.4]{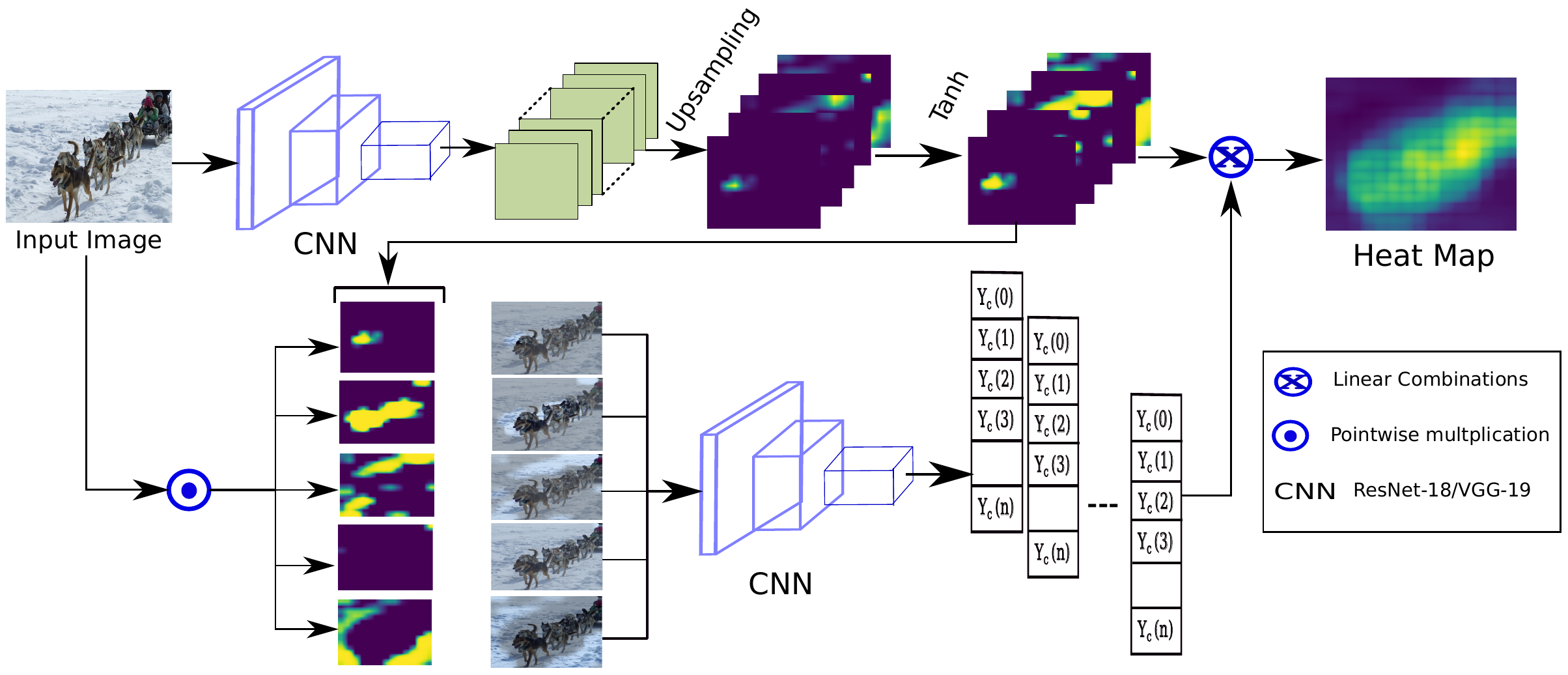}

\caption{Stepwise illustration of our proposed approach: 
  \textbf{Step 1:} Acquiring activation maps from the last layer of the CNN model. 
  \textbf{Step 2:} Upsampling the activation maps to the size of the image and applying the \emph{tanh} function. 
  \textbf{Step 3:} Performing pointwise multiplication of the maps obtained in Step 2 with the original image. 
  \textbf{Step 4:} Passing the resulting images through the CNN model to obtain its scores. 
  \textbf{Step 5:} Calculating the weighted sum of the maps from Step 2 and their corresponding scores to generate the final output.}
\label{fig:prop}

\end{figure*}

In this paper, we propose a simple yet highly effective approach to enhance the interpretability of deep networks. Our strategy aims to capture not only the crucial regions but also the contextual dependencies and connections between them. We enable the model to dynamically attend to important factors and emphasize their importance in decision-making by including attention mechanisms in the saliency map creation process. Users are given access to a more thorough and understandable grasp of the network's thought through the usage of this attention-driven explanation model, which enables them to comprehend why a certain decision was made. This work proposes a novel approach, ScoreCAM++, which leverages simple modifications to the ScoreCAM~\cite{wang2020score} method and significantly boosts the model's performance. The proposed approach changes the normalization function for the activation layer used in ScoreCAM~\cite{wang2020score} and gates the lower-priority values in the activation layer (see Fig. \ref{fig:prop}). Furthermore, we propose a normalization technique adjustment by employing tanh in place of min-max normalization, aiming to amplify the contrast between high and low-priority regions within the activation layer. These simple modifications improve the model's performance and outperform the recent state-of-the-art methods.

We evaluate the effectiveness of our proposed approach on two datasets: the \textit{Cats and Dogs Dataset}~\cite{parkhi2012cats} and the \textit{ImageNet Dataset}~\cite{5206848,su2020does}. Experiments are conducted on two architectures, namely ResNet-18~\cite{he2016deep} and VGG-19~\cite{simonyan2014very}. We showcase the superior interpretability and performance of our approach through a comparison of results with existing explanation methods across different datasets. We present both qualitative and quantitative analyses. The qualitative analysis includes visualizations of the generated saliency maps, highlighting the attention given to relevant features in the input images. In the quantitative analysis, we employ several standard evaluation metrics to measure the performance of our approach compared to the baseline methods. The ablation further validates the advantage of the proposed components in our approach. The key contribution can be summarized as follows:

\begin{itemize}
    \item We propose a simple yet highly effective modification to the ScoreCAM method, which outperforms the existing state-of-the-art methods by a significant margin.
    \item In the proposed approach, we replace the existing normalization of ScoreCAM and apply \emph{tanh} before the Hadamard multiplication with the input image, aiming to amplify the contrast between high and low-priority regions. For the same reason, we apply \emph{tanh} to the upsampled activations before multiplying the channel-wise increase of the confidence score to it.
    \item Extensive experiments over the ResNet-18 and VGG-19 architectures for the two datasets demonstrate a significant improvement for all standard metrics. Additionally, the ablation study emphasizes the importance of the proposed components.
\end{itemize}

\section{Related Work}
Explanation of Deep Neural Networks (DNNs) aims to enhance the interpretability of models for human comprehension of inference logic. One effective strategy to provide explanations involves visualizing specific quantities of interest, such as input feature importance or learned weights, to establish user trust. Spatial convolution, a key component in DNNs, is widely utilized for feature extraction in image and language processing domains. To improve the explication of convolution operations and Convolutional Neural Networks (CNNs), several approaches have gained prominence, including Gradient visualization \cite{simonyan2014very}, Perturbation \cite{ribeiro2016should}, and Class Activation Map (CAM) \cite{zhou2016learning}.

Gradient-based techniques propagate gradients of target classes to the input layer, highlighting regions in the image that strongly influence predictions. Simonyan et al. \cite{simonyan2014very} employ derivatives of target class scores with respect to input images to create saliency maps. Other researchers \cite{zeiler2014visualizing,springenberg2014striving,sundararajan2017axiomatic,adebayo2018local,omeiza2019smooth} manipulate these gradients, enhancing the quality of saliency maps. Nonetheless, these maps often suffer from noise and low quality \cite{omeiza2019smooth}. Perturbation-based methods \cite{ribeiro2016should,petsiuk2018rise,fong2017interpretable,chang2018explaining,dabkowski2017real,wagner2019interpretable} perturb input data to observe model prediction changes. Such approaches often require additional regularizations \cite{fong2017interpretable} and are computationally intensive.

Class Activation Mapping (CAM) \cite{zhou2016learning} identifies discriminative regions by weighted combinations of activation maps from the last convolutional layer before global pooling. While CAM requires a global pooling layer \cite{lin2013network}, Grad-CAM \cite{selvaraju2017grad} extends this technique to diverse CNN architectures without specific architectural constraints. CAM-based explanations \cite{zhou2016learning,selvaraju2017grad,chattopadhay2018grad} visually elucidate single inputs using weighted combinations of activation maps from convolutional layers. Although CAM offers localized visual explanations, it is sensitive to architecture and necessitates a subsequent global pooling layer \cite{lin2013network}. Grad-CAM \cite{selvaraju2017grad} and its variants, such as Grad-CAM++ \cite{chattopadhay2018grad}, aim to generalize CAM, gaining wide acceptance in the research community. XGradCAM \cite{fu2020axiom} introduces an approach to the visualization of Class Activation Mapping (CAM) methods by introducing two fundamental axioms, namely Conservation and Sensitivity. Empirical evaluations establish that XGrad-CAM enhances Grad-CAM by ensuring improved conservation and sensitivity properties. LayerCAM \cite{jiang2021layercam} presents a technique that reexamines the intrinsic relationships between feature maps and their corresponding gradients. By doing so, it achieves the generation of dependable class activation maps across various layers of CNNs. In contrast, ScoreCAM \cite{wang2020score} questions the efficacy of gradients for generalizing CAM and introduces a post-hoc method named ScoreCAM. This method encodes the importance of activation maps through global feature contributions instead of local sensitivity measurements, addressing the limitations of gradient-based CAM variations.

Augmented Grad-CAM++ \cite{electronics12234846} introduces a saliency map generation approach that relies on image geometry augmentation and super-resolution, termed augmented high-order gradient weighting class activation mapping. Statistic-CAM \cite{10295912} suggests a method to generate class activation maps for CNN models in a gradient-free and computationally efficient manner. CAMF \cite{10288391} proposes a learnable fusion rule for infrared and visible image fusion, leveraging class activation mapping. AD-CAM \cite{10304247} presents a technique that exploits the connection between activation maps and network predictions, incorporating temporary masking of specific feature maps.
In this work, we revisit ScoreCAM \cite{wang2020score} and propose ScoreCAM++, a simple yet highly effective variant. ScoreCAM++ improves results by modifying the normalization function of the activation layer in ScoreCAM, effectively gating lower-priority values within the activation layer.

\section{Background}
We define a Convolution Neural Network architecture as a function $y = f(X)$, where $X$ is a tensor, $f$ is the CNN model, and $y$ is the probability distribution of the various classes outputted by the model. $y^c$ defines the probability of the class `$c$'. For a layer $l$, in the CNN, $A_l$ defines the activations function of the layer, and $A^k_l$ defines the activations of the $k^{th}$ channel.

The Grad-CAM~\cite{selvaraju2017grad} is a popular visualization method, which uses the gradient of the last convolutional layer utilizing a backward pass. It is defined as: 
\begin{equation}
L_{Grad-CAM}^c = ReLU(\sum_k \alpha_k^c \cdot A^k_{l})   
\end{equation}

moreover, $\alpha_k^c$ is defined as:
\begin{equation}
\alpha_k^c = GP\left(\frac{{\partial Y^c}}{{\partial A\textsuperscript{k}_{l}}}\right)
\end{equation}
where GP(.) denotes the global pooling operation. There are numerous variants of Grad-CAM, such as Grad-CAM++, which uses various combinations of gradients to obtain the weights multiplied with the activation layers of the last convolutional layer.

The gradient-based methods are not robust and face significant challenges. For example, gradients may vanish due to saturation in the Sigmoid and the flat part of the ReLU activation functions, thus leading to noisy saliency map generation. Additionally, there is an issue of false confidence; using the average pool of the gradients, in some cases, leads to false confidence and wrongly gives high priorities to less contributing activation layers. To address these highly challenging issues, ScoreCAM was introduced, which employed a non-gradient-based approach to generate the saliency map.

However, the normalization function of the ScoreCAM method does not provide the discriminative score at the activation layer that can easily identify the low and high-priority regions. A function with good differentiating ability between the higher and lower priority values in the activation layer would result in better interpretability and provide a robust and stable saliency map. We achieve the same by modifying the normalization function to have more discriminative power. In the following section, we will describe the proposed ScoreCAM++ approach.

\begin{table*}[t]
\centering
\caption{Results obtained using ResNet-18 architecture over the Cat and Dog Dataset for various methods. AugGradCAM++ refers to Augmented GradCAM++. Average Drop\%: lower is better. Increase in Confidence and Win\%: higher is better.}

\resizebox{\textwidth}{!}{
\begin{tabular}{c|c|c|c|c|c|c|c}
\hline
& GradCAM & GradCAM++ & XGradCAM & LayerCAM & ScoreCAM  & AugGradCAM++ & ScoreCAM++ \\ \hline
Average Drop \%        & 8.62    & 8.55      & 8.62     & 8.48     & 6.89  & 9.22   & \textbf{3.29}                    \\ \hline
Increase in Confidence & 34.95   & 33.15     & 34.85    & 33.25    & 39.25    & 28.00           & \textbf{50.25}        \\ \hline
Win \%                 & 10.05   & 3.20      & 4.60     & 1.50     & 14.75   & 7.70     & \textbf{58.20}       \\ \hline
\end{tabular}}
\label{tab:res_dogcat}

\end{table*}

\begin{table*}[t]
\centering
\caption{Results obtained using VGG-19 architecture over the Cat and Dog Dataset for various methods. AugGradCAM++ refers to Augmented GradCAM++. Average Drop\%: lower is better. Increase in Confidence and Win\%: higher is better.}

\resizebox{\textwidth}{!}{
\begin{tabular}{c|c|c|c|c|c|c|c}
\hline
                       & GradCAM & GradCAM++ & XGradCAM & LayerCAM & ScoreCAM & AugGradCAM++ & ScoreCAM++ \\ \hline
Average Drop \%        & 8.92    & 7.22      & 7.08     & 7.64     & 7.33   & 7.59             & \textbf{3.24}    \\ \hline
Increase in Confidence & 48.30   & 47.05     & 52.70    & 46.00    & 47.05     & 42.45     & \textbf{60.95}               \\ \hline
Win \%                 & 12.85   & 6.15      & 16.65    & 6.40     & 13.50       & 7.15    & \textbf{37.30}               \\ \hline
\end{tabular}}
\label{tab:vgg_dogcat}

\end{table*}

\begin{figure*}[t]
\small
\begin{center}
\addtolength{\tabcolsep}{-1.5mm}
\begin{tabular}{cccccccc}
Original & GradCAM & GradCAM++ &XGradCAM & LayerCAM & ScoreCAM&AugGradCAM++  & ScoreCAM++\\
\includegraphics[height=2.0cm,width=2.0cm]{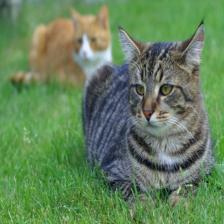}
	&
\includegraphics[height=2.0cm,width=2.0cm]{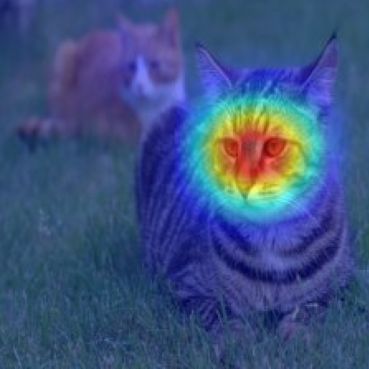}
	&
\includegraphics[height=2.0cm,width=2.0cm]{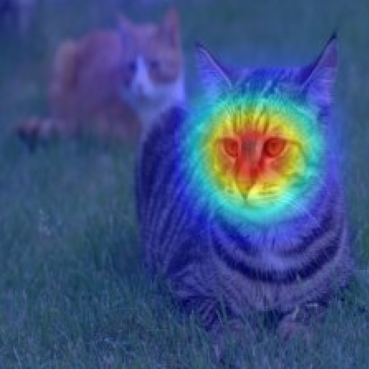}	
	&
\includegraphics[height=2.0cm,width=2.0cm]{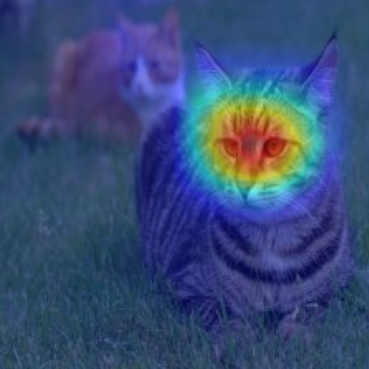}
 &
\includegraphics[height=2.0cm,width=2.0cm]{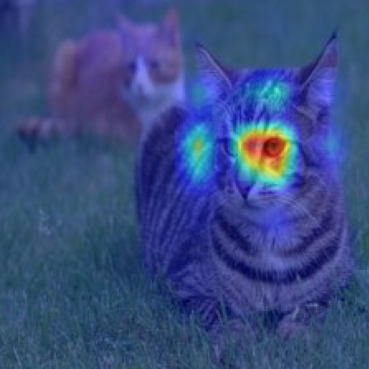}	
    &
\includegraphics[height=2.0cm,width=2.0cm]{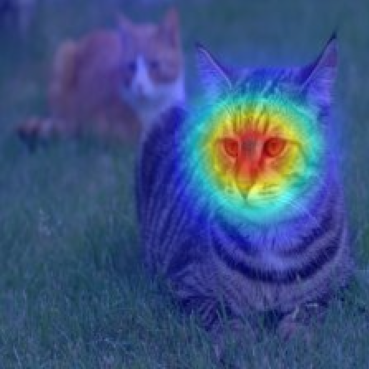}
	&
\includegraphics[height=2.0cm,width=2.0cm]{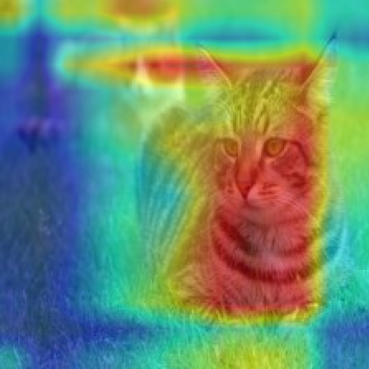}
	&
\includegraphics[height=2.0cm,width=2.0cm]{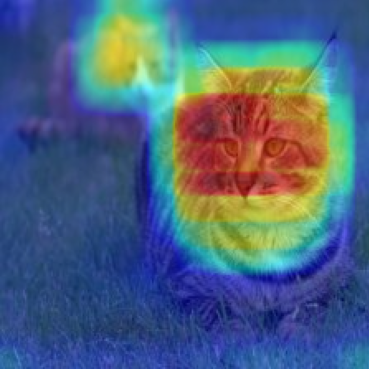}
\\ 
\includegraphics[height=2.0cm,width=2.0cm]{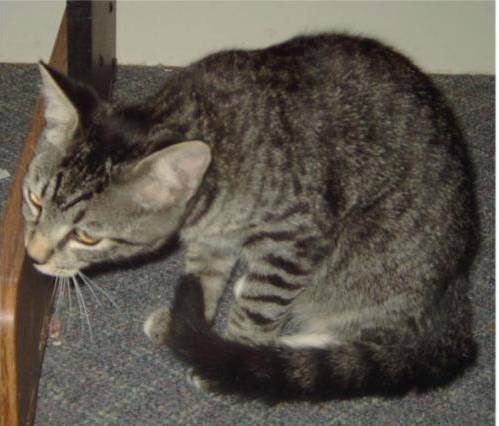}
	&
\includegraphics[height=2.0cm,width=2.0cm]{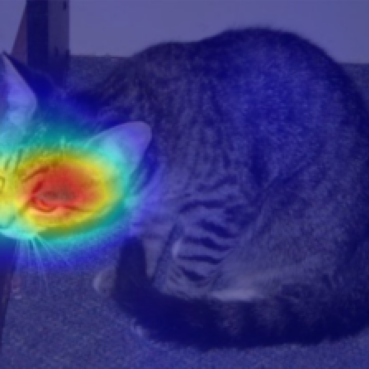}
	&
\includegraphics[height=2.0cm,width=2.0cm]{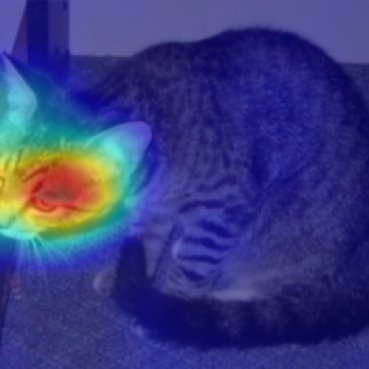}	
	&
\includegraphics[height=2.0cm,width=2.0cm]{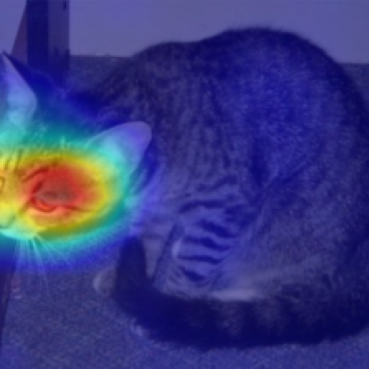}	
        &
\includegraphics[height=2.0cm,width=2.0cm]{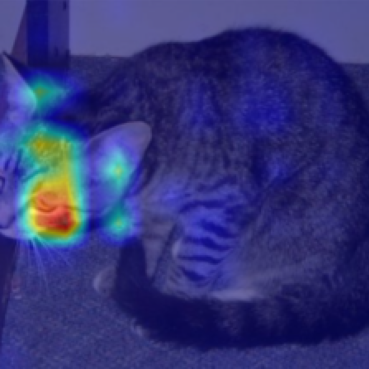}
	&
\includegraphics[height=2.0cm,width=2.0cm]{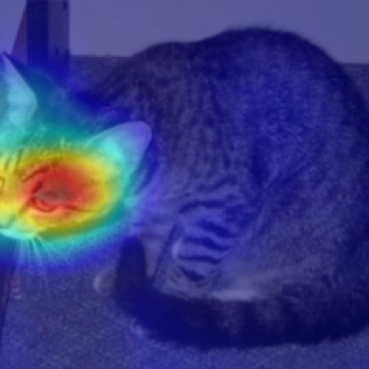}
	&
\includegraphics[height=2.0cm,width=2.0cm]{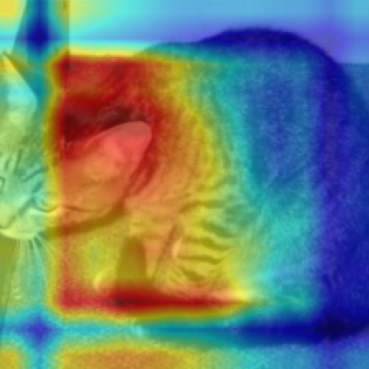}
	&
\includegraphics[height=2.0cm,width=2.0cm]{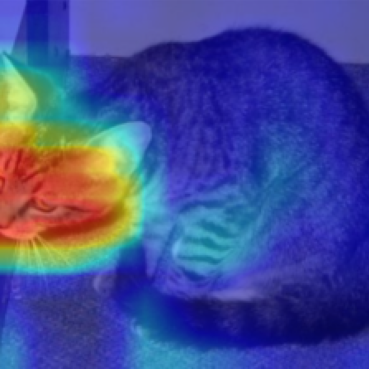}
\\
\includegraphics[height=2.0cm,width=2.0cm]{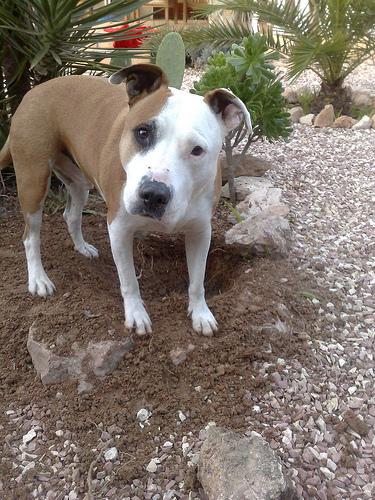}
	&
\includegraphics[height=2.0cm,width=2.0cm]{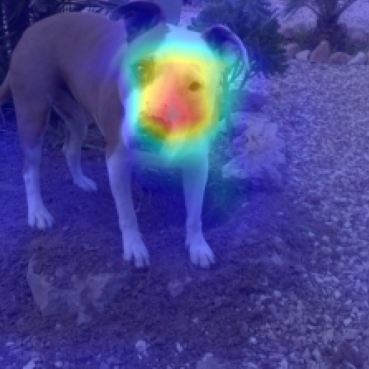}
	&
\includegraphics[height=2.0cm,width=2.0cm]{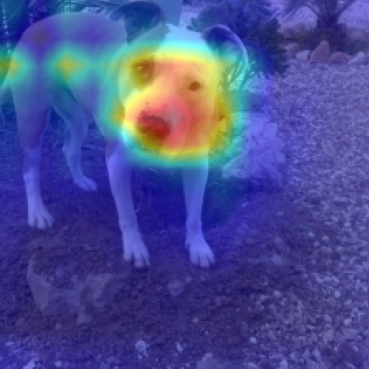}	
	&
\includegraphics[height=2.0cm,width=2.0cm]{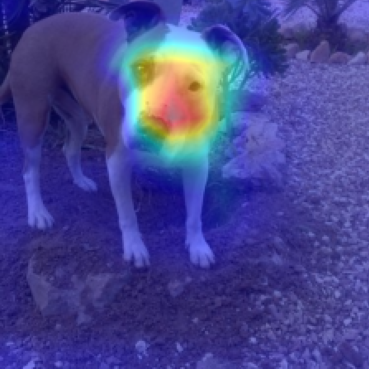}	
        &
\includegraphics[height=2.0cm,width=2.0cm]{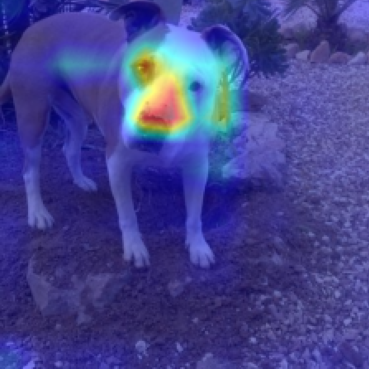}
	&
\includegraphics[height=2.0cm,width=2.0cm]{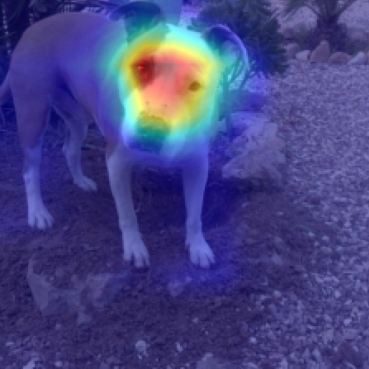}
	&
\includegraphics[height=2.0cm,width=2.0cm]{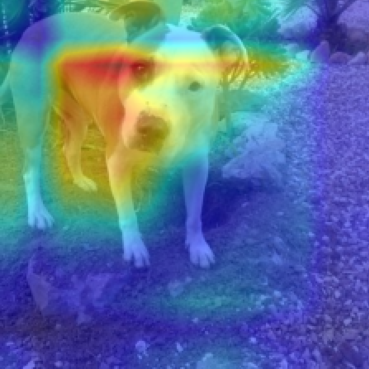}
	&
\includegraphics[height=2.0cm,width=2.0cm]{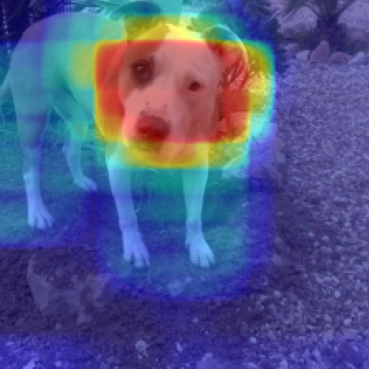}
\\ 
\end{tabular}
\end{center} 

\caption{Qualitative results obtained using VGG-19 architecture over the Cat and Dog Dataset for various methods. AugGradCAM++ refers to Augmented GradCAM++.}

\label{fig:dog_cat_vgg}
\end{figure*}

\section{Proposed Approach: ScoreCAM++}

To address the issues in the gradient-based methods discussed in the previous section, ScoreCAM was introduced. ScoreCAM presents a non-gradient-based technique for generating class activation maps, based on the assumption that the activation layers derived from the final convolutional layer encapsulate essential information for constructing saliency maps. This involves element-wise multiplication between an upsampled activation layer extracted from the last convolutional layer and the original input image. The generated image aims to emphasize the regions highlighted by the activation layer.

The image is propagated through the CNN model to obtain a confidence score for the target class. The score's magnitude is influenced by the activation layer used for multiplication, signifying the relative importance of the chosen layer. This process is replicated for each activation layer produced by the last convolutional layers. The cumulative effect of these layers contributes to the ultimate saliency map. The ScoreCAM can be formulated as:
\begin{equation}
\label{eq:scorecam}
L_{ScoreCAM}^c = ReLU(\sum_k \alpha_k^c \cdot A^k_{l})   
\end{equation}
where, $\alpha_k^c$ is given as:
\begin{equation}
\alpha_k^c = C(A^k_{l})  
\end{equation}
C(.) is the channel-wise increase of confidence score (CIC Score), and it is defined as:
\begin{equation}
C(A^k_{l}) = {f}(X \circ H^k_l) - f(X)
\end{equation}
where,
\begin{equation}
H^{k}_l = s(Up(A^k_{l}))
\end{equation}
and $\circ$ refers to the Hadamard Multiplication, $s(.)$ is the normalization function and $Up$ represents the upsampling operation. In ScoreCAM, normalized upsampled activation layers undergo Hadamard multiplication with the input image, thus standardizing their values using the function $s(.)$ as defined in Eq.~\ref{equ:normalize}.
\begin{equation}
s(A^k_{l}) = \frac{A^k_{l}-min(A^k_{l})}{ min(A^k_{l})-max(A^k_{l})}
\label{equ:normalize}
\end{equation}
The utilization of min-max normalization within ScoreCAM has some limitations; using min-max normalization can hinder the highlighting of high-priority regions in the activation layer, making the model less interpretable. Improving the interpretability of the model requires addressing this issue. Furthermore, the range of $[0,1]$ yielded by min-max normalization is not always optimal. Negative activation values, indicating the level of inhibition or suppression of a particular feature in the input, offer valuable insights into aspects that the model ignores or deems irrelevant to its predictions. Examining these negative activations can uncover factors influencing the model's decision-making. To overcome these limitations, we propose ScoreCAM++, which demonstrates significant improvement over the ScoreCAM method.

In our proposed approach, we introduce a modification by employing an activation function during the scaling process. Our aim is to amplify high-priority activation values while diminishing lower ones. To achieve this, we employ the \emph{tanh} activation function (see Fig.~\ref{fig:prop}). The \emph{tanh} activation function can be defined as:

\begin{equation}
        tanh = \frac{e^x - e^{-x}}{e^x+e^{-x}}
        \label{equ:tanh_act}
\end{equation}

The \emph{tanh} activation function scales activation values in the range of $[-1,1]$, employing tanh on the upsampled activation layer before Hadamard multiplication with the input image causes the upsampled activation layer's values to converge towards 1 for higher priority activations, and towards 0 or -1 for lower or negative priority ones. Our approach establishes a clear distinction between high and low-priority values. As a result, we get the resulting confidence score for the class based on the information that is most important for the activation layer. Furthermore, we apply the \emph{tanh} function to the upsampled activation layers, following the same principle of emphasizing regions with higher priority within these layers. As a result, when aggregating the weighted sum of these activation layers, the resulting saliency map becomes skewed towards these elevated priority regions. This bias significantly enhances the comprehensibility of the predictions, leading to a more refined interpretation.

Thus, for a model $f$ with a convolutional layer 
$l$ and a specific class of interest $c$, the ScoreCAM++ can be defined as follows:
\begin{equation}
\label{eq:scorecam++}
L_{ScoreCAM++}^c = ReLU(\sum_k \alpha_k^c \cdot tanh(A^k_{l}))   
\end{equation}

Where,
\begin{equation}
\begin{aligned}
\alpha_k^c = C(A^k_{l}),\\
C(A^k_l) = f(X \circ H^k_{l}) - f(X)
\end{aligned}
\end{equation}
$H^k_{l}$ is given as:
\begin{equation}
H^k_{l} = s(Up(A^k_l))
\end{equation}
and $s(.)$ is defined as:
\begin{equation}
\label{eq:our_normalize}
s(A^k_l) = tanh(A^k_l)
\end{equation}

In comparison to the original ScoreCAM in Eq.~\ref{eq:scorecam}, which does not use any activation map, our approach (see Eq.~\ref{eq:scorecam++}) incorporates the \emph{tanh} activation to discriminate the activation map across a broader range. The modified definition of the normalization function, from Eq.~\ref{equ:normalize} to Eq.~\ref{eq:our_normalize}, helps suppress the lower-priority regions and directs attention to the most relevant regions of the activation map.

\begin{table*}[t]
\centering
\caption{Results obtained using ResNet-18 architecture over ImageNet Dataset for various methods. AugGradCAM++ refers to Augmented GradCAM++. Average Drop\%: lower is better. Increase in Confidence and Win\%: higher is better.}

\resizebox{\textwidth}{!}{
\begin{tabular}{c|c|c|c|c|c|c|c}
\hline
                       & GradCAM & GradCAM++ & XGradCAM & LayerCAM & ScoreCAM & AugGradCAM++  & ScoreCAM++\\ \hline
Average Drop \%        & 9.77    & 9.98      & 9.75     & 9.87     & 8.48   & 10.46  & \textbf{4.17}                  \\ \hline
Increase in Confidence & 28.45   & 26.30     & 28.60    & 26.65    & 31.15  & 22.20   & \textbf{43.75}                 \\ \hline
Win \%                 & 5.65    & 2.15      & 9.00     & 1.70     & 9.50   & 9.85   & \textbf{62.15}         \\ \hline
\end{tabular}}
\label{tab:res_imagenet}

\end{table*}

\begin{figure*}[t]
\small
\begin{center}
\addtolength{\tabcolsep}{-1.5mm}
\begin{tabular}{cccccccc}
Original & GradCAM & GradCAM++ & XGradCAM & LayerCAM & ScoreCAM &AugGradCAM++ & ScoreCAM++\\
\includegraphics[height=2.0cm,width=2.0cm]{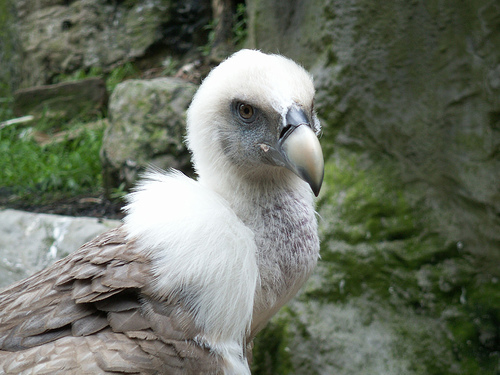}
	&
\includegraphics[height=2.0cm,width=2.0cm]{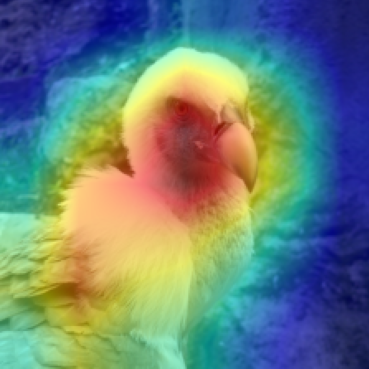}
	&
\includegraphics[height=2.0cm,width=2.0cm]{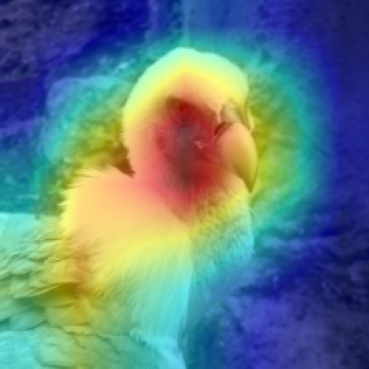}	
	&
\includegraphics[height=2.0cm,width=2.0cm]{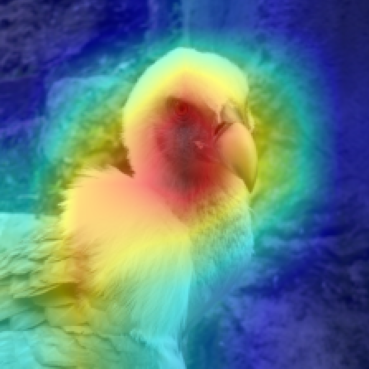}
 &
\includegraphics[height=2.0cm,width=2.0cm]{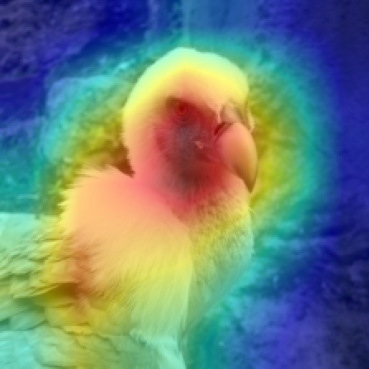}	
    &
\includegraphics[height=2.0cm,width=2.0cm]{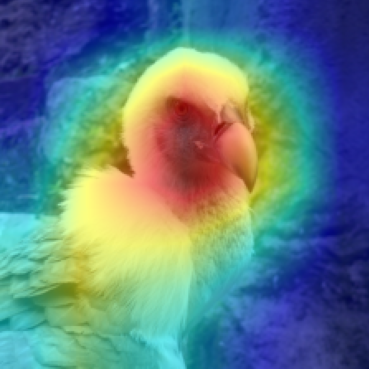}
	&
\includegraphics[height=2.0cm,width=2.0cm]{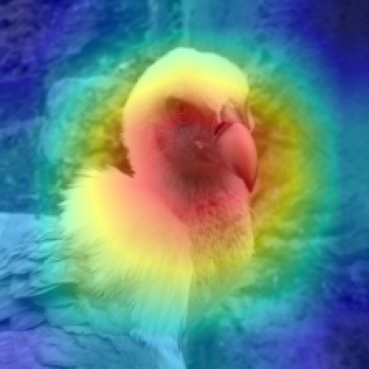}
	&
\includegraphics[height=2.0cm,width=2.0cm]{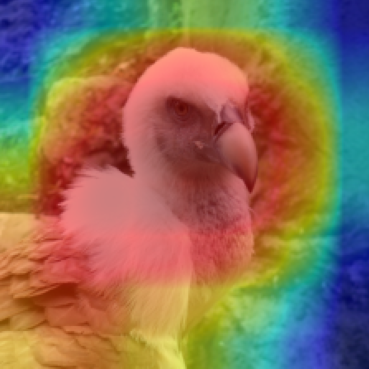}
\\ 
\includegraphics[height=2.0cm,width=2.0cm]{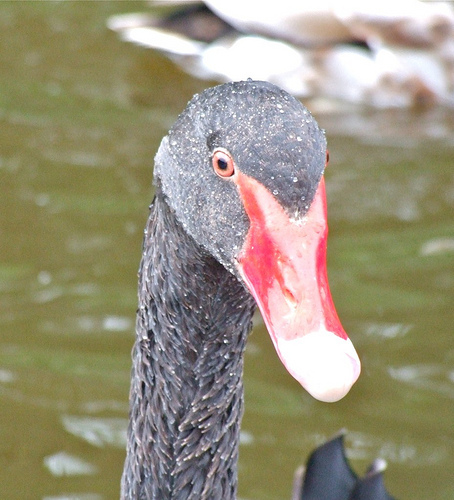}
	&
\includegraphics[height=2.0cm,width=2.0cm]{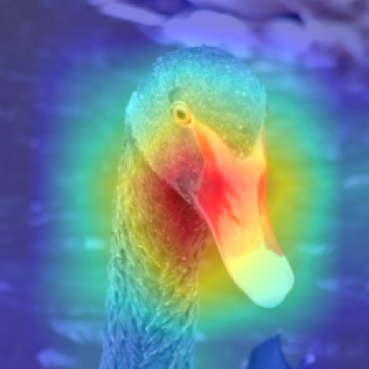}
	&
\includegraphics[height=2.0cm,width=2.0cm]{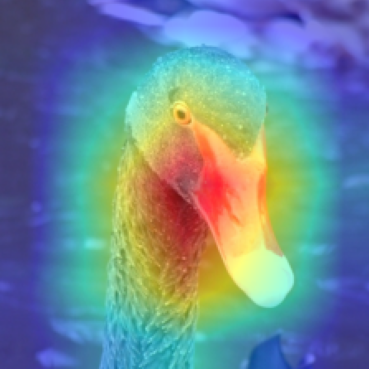}	
	&
\includegraphics[height=2.0cm,width=2.0cm]{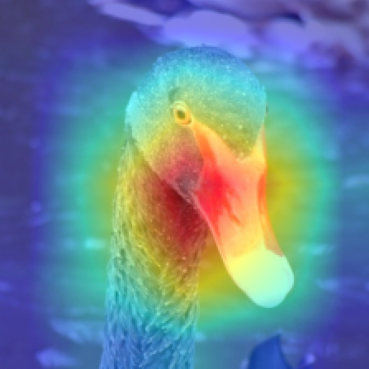}		
    &
\includegraphics[height=2.0cm,width=2.0cm]{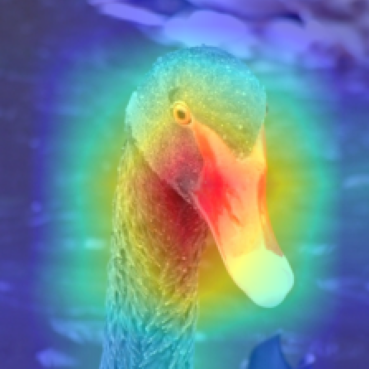}
	&
\includegraphics[height=2.0cm,width=2.0cm]{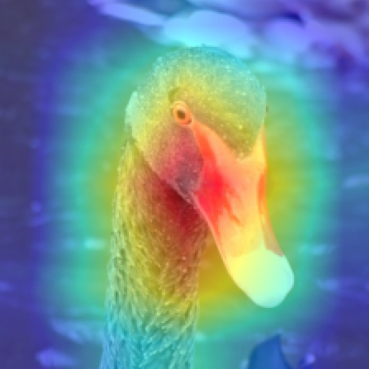}
	&
\includegraphics[height=2.0cm,width=2.0cm]{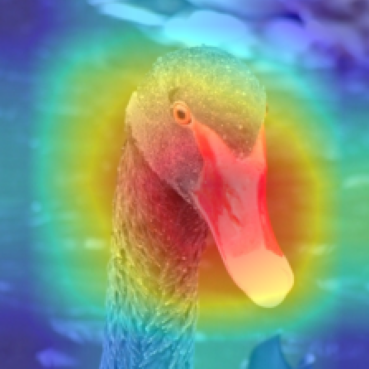}
	&
\includegraphics[height=2.0cm,width=2.0cm]{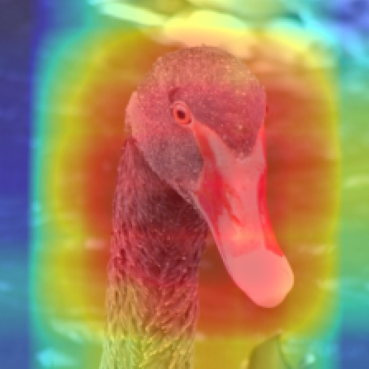}
\\
\includegraphics[height=2.0cm,width=2.0cm]{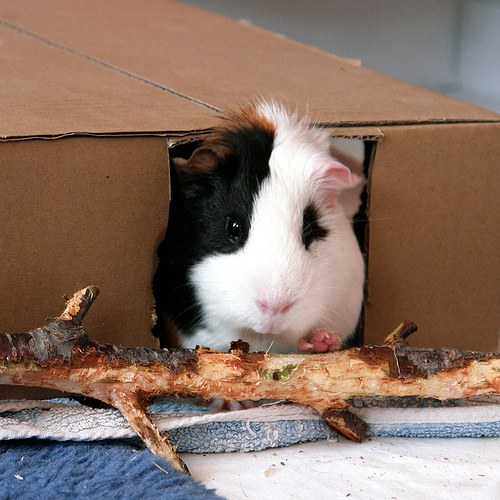}
	&
\includegraphics[height=2.0cm,width=2.0cm]{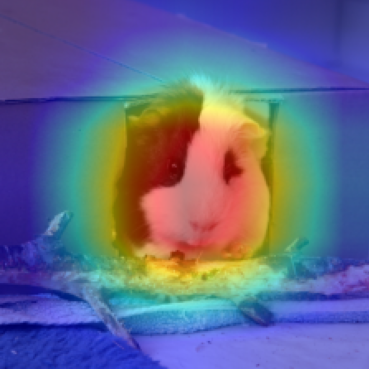}
	&
\includegraphics[height=2.0cm,width=2.0cm]{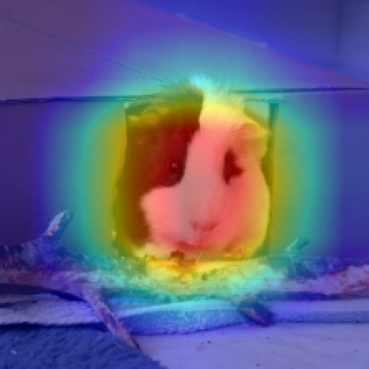}	
	&
\includegraphics[height=2.0cm,width=2.0cm]{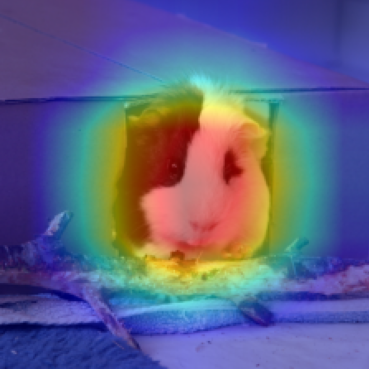}	
        &
\includegraphics[height=2.0cm,width=2.0cm]{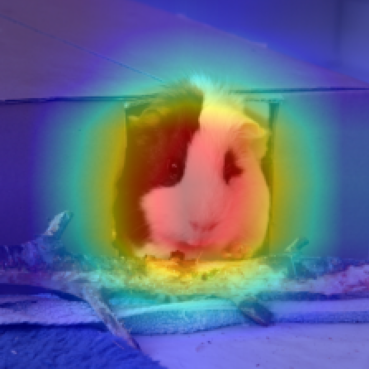}
	&
\includegraphics[height=2.0cm,width=2.0cm]{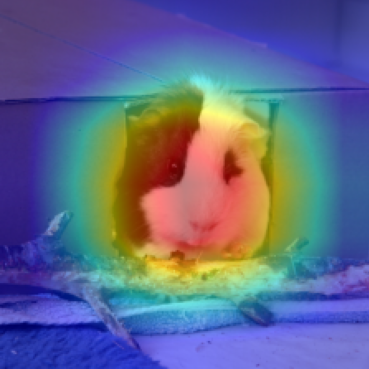}
	&
\includegraphics[height=2.0cm,width=2.0cm]{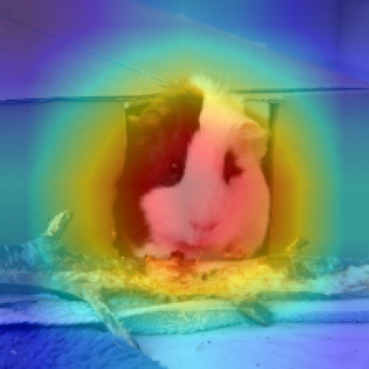}
	&
\includegraphics[height=2.0cm,width=2.0cm]{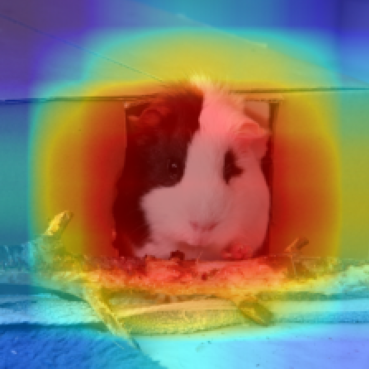}
\\ 
\end{tabular}
\end{center} 

\caption{Qualitative results obtained using ResNet-18 architecture over the ImageNet Dataset for various methods. It is observed that ScoreCAM++ provides the best visual explanations compared to baseline methods. AugGradCAM++ refers to Augmented GradCAM++.}
 \label{fig:resnet_imagenet}
 
\end{figure*}

\begin{figure*}[t]
\small
\begin{center}
\addtolength{\tabcolsep}{-1.5mm}
\begin{tabular}{cccccccc}
Original & GradCAM & GradCAM++ &XGradCAM & LayerCAM & ScoreCAM  &AugGradCAM++& ScoreCAM++\\
\includegraphics[height=2.0cm,width=2.0cm]{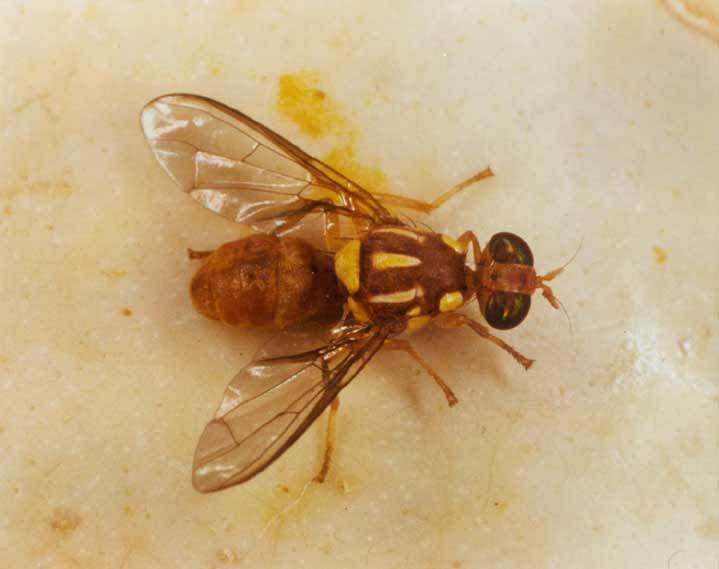}
	&
\includegraphics[height=2.0cm,width=2.0cm]{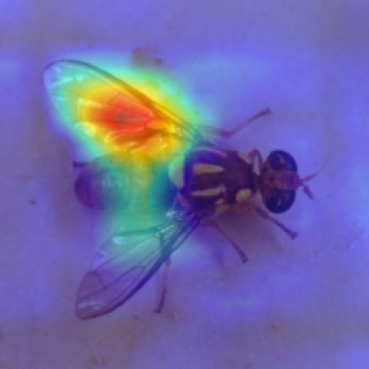}
	&
\includegraphics[height=2.0cm,width=2.0cm]{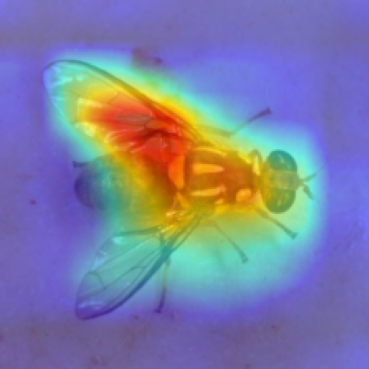}	
	&
\includegraphics[height=2.0cm,width=2.0cm]{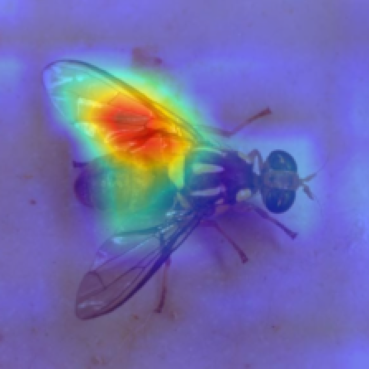}	
        &
\includegraphics[height=2.0cm,width=2.0cm]{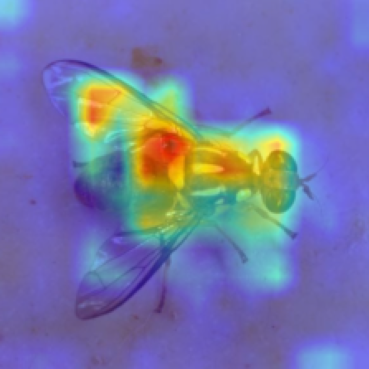}
	&
\includegraphics[height=2.0cm,width=2.0cm]{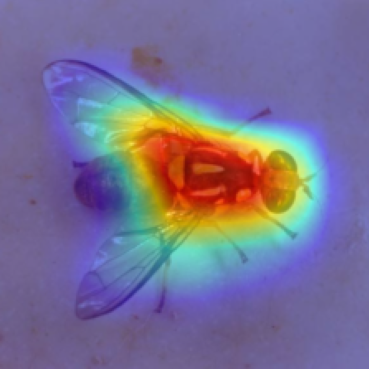}
	&
\includegraphics[height=2.0cm,width=2.0cm]{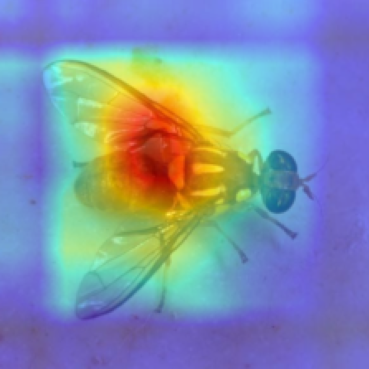}
	&
\includegraphics[height=2.0cm,width=2.0cm]{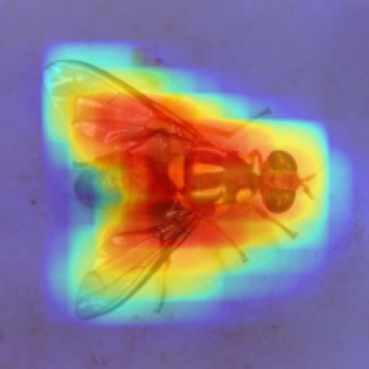}
\\
\includegraphics[height=2.0cm,width=2.0cm]{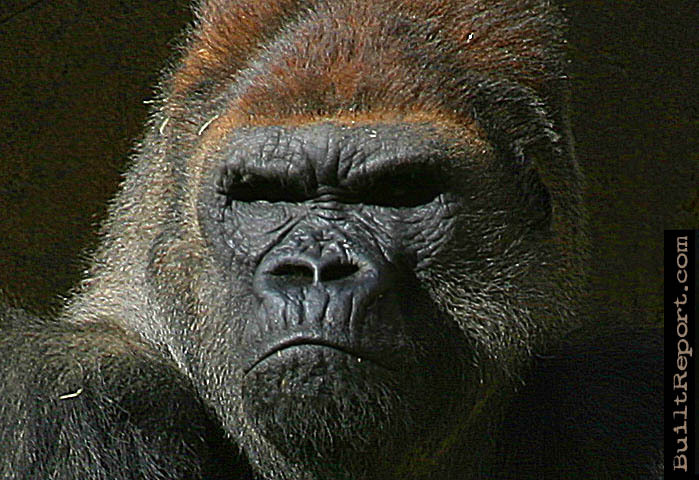}
	&
\includegraphics[height=2.0cm,width=2.0cm]{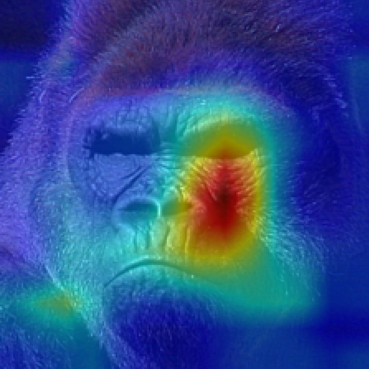}
	&
\includegraphics[height=2.0cm,width=2.0cm]{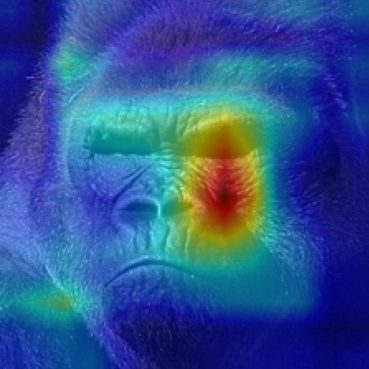}	
	&
\includegraphics[height=2.0cm,width=2.0cm]{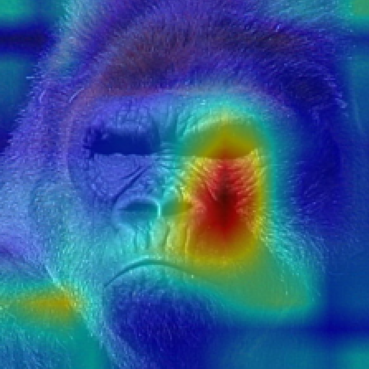}	
        &
\includegraphics[height=2.0cm,width=2.0cm]{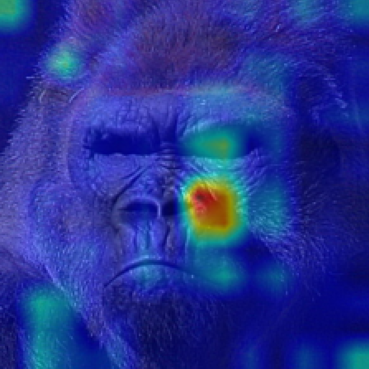}
	&
\includegraphics[height=2.0cm,width=2.0cm]{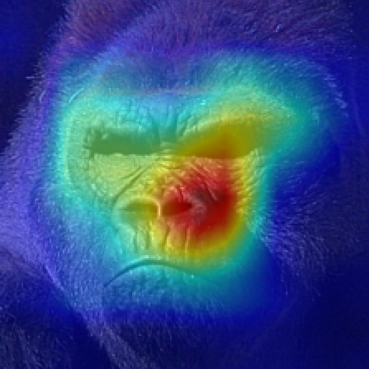}
	&
\includegraphics[height=2.0cm,width=2.0cm]{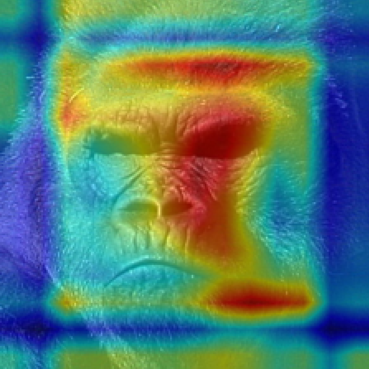}
	&
\includegraphics[height=2.0cm,width=2.0cm]{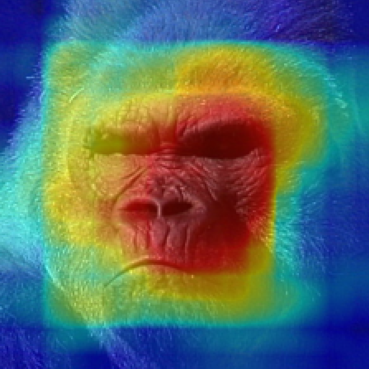}
\\
\includegraphics[height=2.0cm,width=2.0cm]{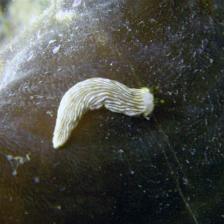}
	&
\includegraphics[height=2.0cm,width=2.0cm]{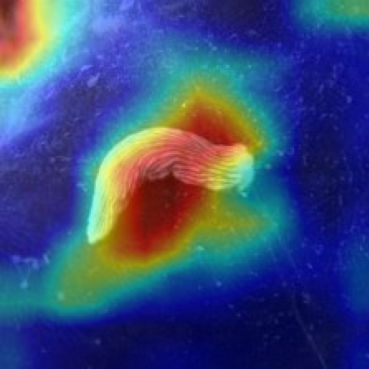}
	&
\includegraphics[height=2.0cm,width=2.0cm]{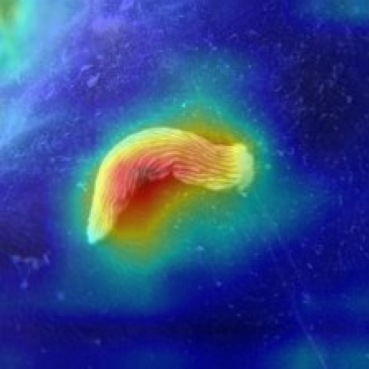}	
	&
\includegraphics[height=2.0cm,width=2.0cm]{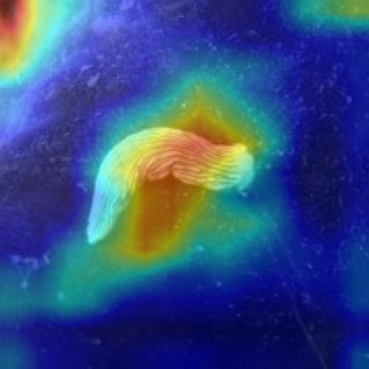}	
        &
\includegraphics[height=2.0cm,width=2.0cm]{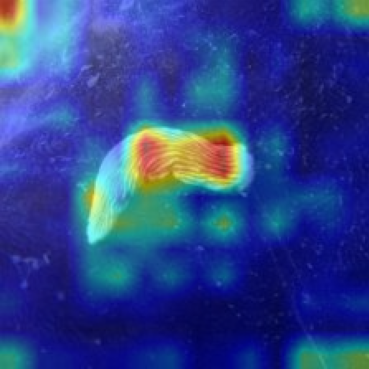}
	&
\includegraphics[height=2.0cm,width=2.0cm]{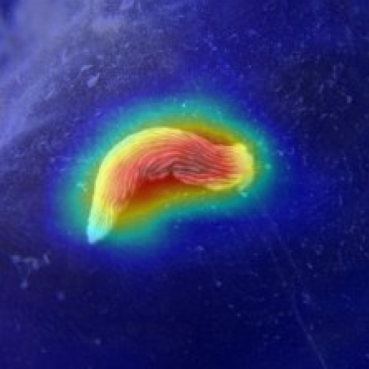}
	&
\includegraphics[height=2.0cm,width=2.0cm]{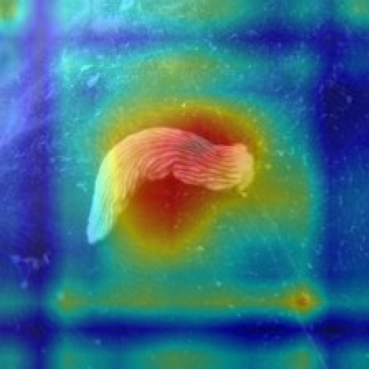}
	&
\includegraphics[height=2.0cm,width=2.0cm]{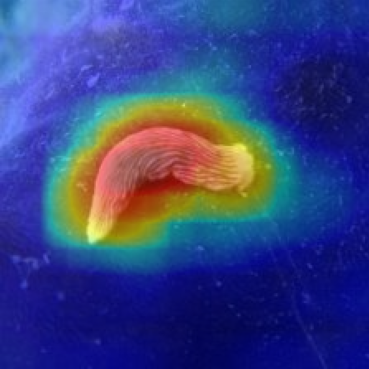}
\end{tabular}
\end{center} 

\caption{Qualitative results obtained using VGG-19 architecture over the ImageNet Dataset for various methods. It is observed that ScoreCAM++ provides the best visual explanations compared to baseline methods. AugGradCAM++ refers to Augmented GradCAM++.}
 \label{fig:vgg_imagenet}

\end{figure*}

\begin{table*}[t]
\centering
\caption{Results obtained using VGG-19 architecture over ImageNet Dataset for various methods. AugGradCAM++ refers to Augmented GradCAM++. Average Drop\%: lower is better. Increase in Confidence and Win\%: higher is better.}

\resizebox{\textwidth}{!}{
\begin{tabular}{c|c|c|c|c|c|c|c}
\hline
                       & GradCAM & GradCAM++ & XGradCAM & LayerCAM & ScoreCAM  & AugGradCAM++ & ScoreCAM++\\ \hline
Average Drop \%        & 12.51   & 13.87     & 11.43    & 13.17    & 12.75     & 12.93  & \textbf{7.28}                   \\ \hline
Increase in Confidence & 28.85   & 23.95     & 30.95    & 24.40    & 26.75        & 21.40  & \textbf{38.25}               \\ \hline
Win \%                 & 12.50   & 4.25      & 17.35    & 5.05     & 9.05     & 10.65  & \textbf{41.15}              \\ \hline
\end{tabular}}
\label{tab:vgg_imagenet}

\end{table*}

\section{Experimental Results}

We assess the performance of our approach using three metrics (detailed descriptions of the evaluation metrics are provided in the Appendix section~\ref{sec:metrics}): Average Drop Percentage, Increase in Confidence (also recognized as Average Increase), and Win Percentage. These metrics are standard evaluation metrics, which have been utilized in several prior works, including ScoreCAM \cite{wang2020score} and GradCAM++ \cite{chattopadhay2018grad}.  A comparative analysis is conducted across several widely used interpretable techniques, namely, GradCAM \cite{selvaraju2017grad}, GradCAM++ \cite{chattopadhay2018grad}, XGradCAM \cite{fu2020axiom}, LayerCAM \cite{jiang2021layercam}, Augmented Grad-CAM++ \cite{electronics12234846}, and ScoreCAM \cite{wang2020score}, along with the proposed ScoreCAM++. The aforementioned metrics are used to facilitate the comparison among these methods. Qualitative and quantitative experiments are conducted on two distinct datasets: \emph{ImageNet} dataset~\cite{krizhevsky2012imagenet,su2020does} and \emph{Dogs and Cats} dataset~\cite{parkhi2012cats}. The evaluations are conducted using pretrained VGG-19 \cite{simonyan2014very} and ResNet-18 \cite{he2016deep} architectures. We analyze the reliability of explanations generated by the proposed ScoreCAM++ in the context of image classification tasks. For a fair comparison, we follow a similar setting as ScoreCAM, and all baseline methods are evaluated under the same conditions. The reported results for ImageNet are average over a random sample of 2000 images from the validation dataset, while on the Dogs and Cats dataset, 4000 images are chosen randomly from a pool of 12,500 test images. The Cat and Dog training dataset, consisting of 25,000 images of cats and dogs, is used to fine-tune the VGG-19 and ResNet-18 networks, which were originally trained on ImageNet. Details regarding the implementation are provided in the Appendix section ~\ref{sec:implem}.

\subsection{Results on Dog and Cat Dataset}

Table~\ref{tab:res_dogcat} presents the results for the ResNet-18 model on the Cat and Dog dataset, while Table~\ref{tab:vgg_dogcat} shows the results for the VGG-19 architecture. ScoreCAM++ outperforms ScoreCAM and all other compared methods on all three metrics. A lower average drop percentage, a higher increase in confidence, and a higher win percentage demonstrate the efficacy of the proposed method in generating visual explanations. As shown in Table~\ref{tab:res_dogcat}, ScoreCAM++ with the ResNet-18 model exhibits an average drop of 3.29\%, an increase in confidence of 50.25\%, and a win percentage of 58.20\%, respectively. Similarly, Table~\ref{tab:vgg_dogcat} presents the outcomes of ScoreCAM++ when employed with the VGG-19 network, showing an average drop of 3.24\%, an increase in confidence by 60.95\%, and a win percentage of 37.30\%. This highlights the ability of ScoreCAM++ to adeptly emphasize the most prominent regions of the target object.

A comparison of visual explanations generated by various explainable methods, along with our proposed ScoreCAM++, is shown in Fig.~\ref{fig:dog_cat_vgg}. It is evident from Fig.~\ref{fig:dog_cat_vgg} that the explanation map generated by ScoreCAM++ accurately reflects the features of the cat or dog. While the explanation maps produced by ScoreCAM highlight the image's distinctive region of interest, they lack the same confidence level observed in ScoreCAM++. As a result, ScoreCAM++ emerges as a trustworthy explainable method for analyzing the complex mechanisms behind a model's decision.

\begin{figure*}[t]
\begin{center}
\addtolength{\tabcolsep}{0.0mm}
\begin{tabular}{cccccc}
Original & ReLU & Sigmoid & Swish & Mish & Tanh  \\
\includegraphics[height=1.8cm,width=1.8cm]{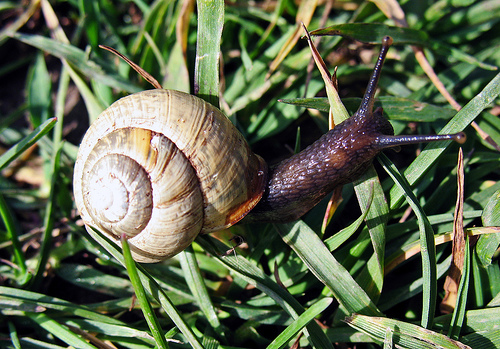}
	&
\includegraphics[height=1.8cm,width=1.8cm]{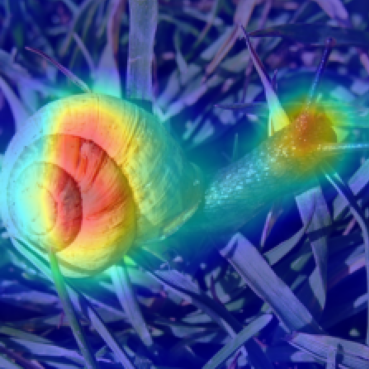}
	&
\includegraphics[height=1.8cm,width=1.8cm]{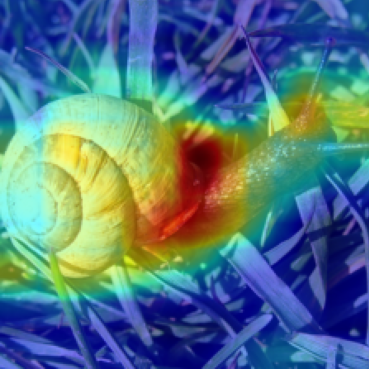}	
	&
\includegraphics[height=1.8cm,width=1.8cm]{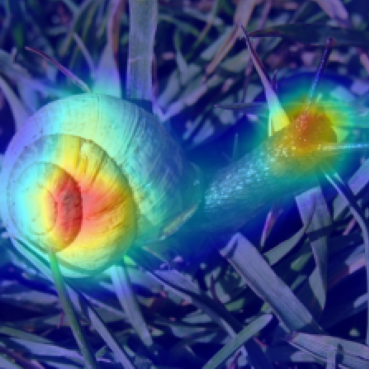}	
    &
\includegraphics[height=1.8cm,width=1.8cm]{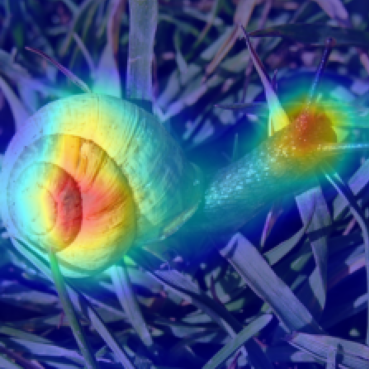}
	&
\includegraphics[height=1.8cm,width=1.8cm]{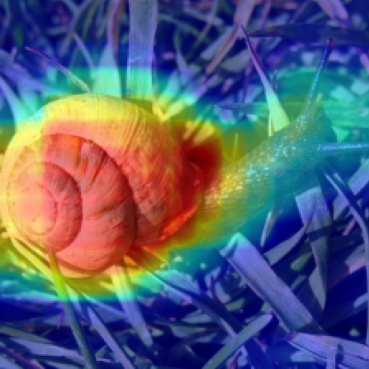}
\\
\includegraphics[height=1.8cm,width=1.8cm]{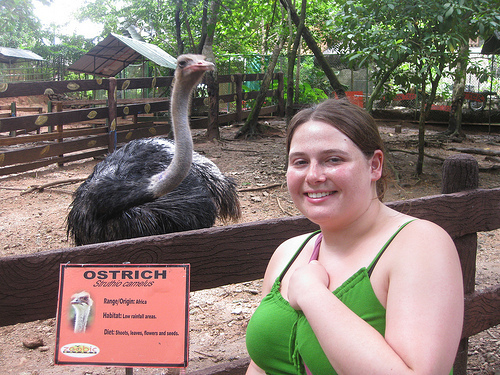}
&
\includegraphics[height=1.8cm,width=1.8cm]{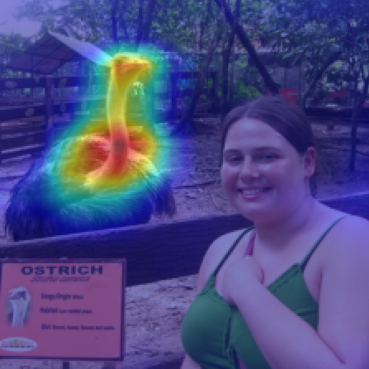}
&
\includegraphics[height=1.8cm,width=1.8cm]{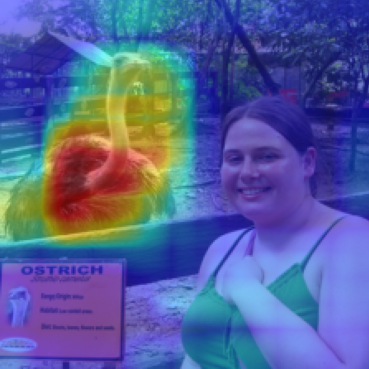}	
	&
\includegraphics[height=1.8cm,width=1.8cm]{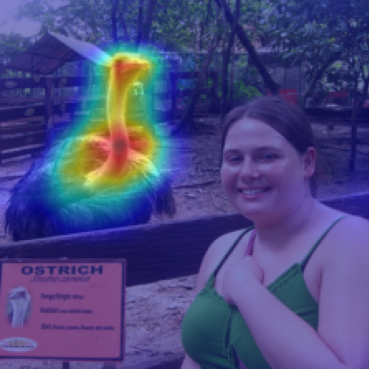}	
        &
\includegraphics[height=1.8cm,width=1.8cm]{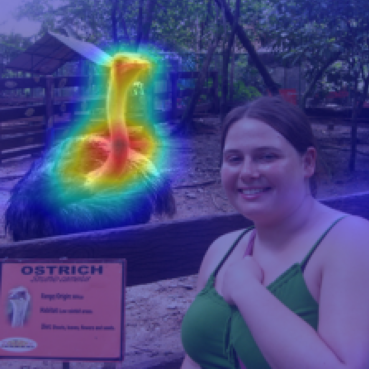}
	&
\includegraphics[height=1.8cm,width=1.8cm]{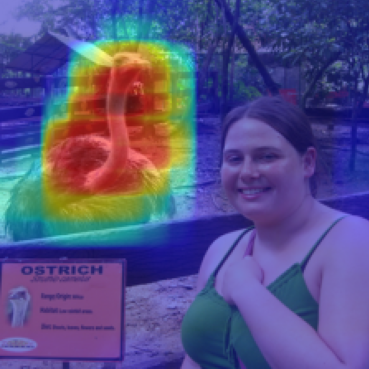}
\\(a) & (b) & (c) & (d) & (e) & (f)\\
\\ 
\end{tabular}
\end{center} 

\caption{Visual explanations generated by ScoreCAM++ using VGG-19 over ImageNet dataset with different types of activation functions.}
 \label{fig:diff_act}

\end{figure*}

\begin{table}[h]
\small
\addtolength{\tabcolsep}{-4.8mm}
\centering
\caption{Ablation experiments using VGG-19 over the ImageNet dataset involving various activation functions in ScoreCAM++. AD stands for \emph{Average Drop}, and IiC stands for \emph{Increase in Confidence}.}

\addtolength{\tabcolsep}{4.5mm}
\begin{tabular}{c|c|c|c|c|c}
\hline
                       & Tanh(.) & ReLU(.) & Sigmoid(.) & Mish(.) & Swish(.) \\ \hline
AD\%         & \textbf{4.27}    & 7.97          &6.04  & 8.71    & 8.97     \\ \hline
IiC & \textbf{43.14}   & 34.79       &34.43     & 32.95   & 32.14    \\ \hline
Win\%                  & \textbf{51.65}   & 15.28       &15.05     & 6.16   & 11.84    \\ \hline
\end{tabular}
\label{tab:diff_act}

\end{table}

\begin{figure}[t]

\centering
\addtolength{\tabcolsep}{0.0mm}
\begin{tabular}{ccc}
Original & ScoreCAM & ScoreCAM++  \\
\includegraphics[height=2.0cm,width=2.0cm]{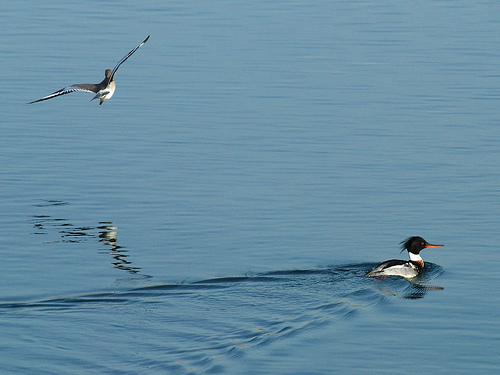}

	&
\includegraphics[height=2.0cm,width=2.0cm]{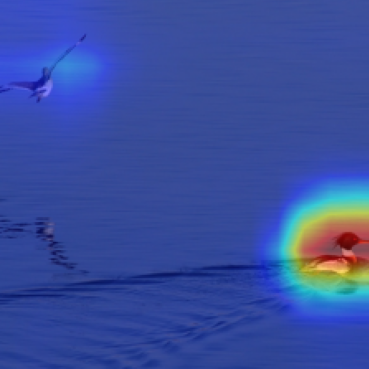}
	&
\includegraphics[height=2.0cm,width=2.0cm]{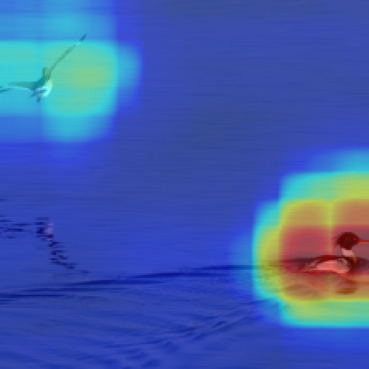}
\\
\includegraphics[height=2.0cm,width=2.0cm]{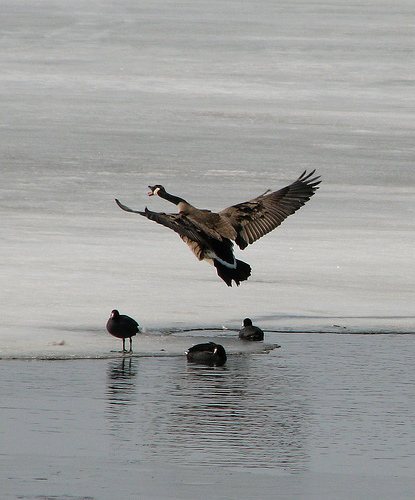}
	&
\includegraphics[height=2.0cm,width=2.0cm]{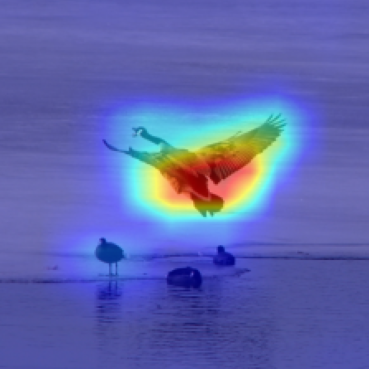}
	&
\includegraphics[height=2.0cm,width=2.0cm]{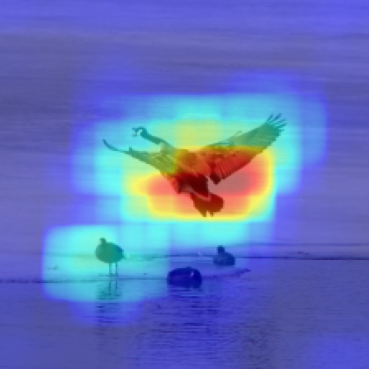}

\\(a) & (b) & (c)\\
\\ 
\end{tabular}

\caption{Visual explanations generated by ScoreCAM++ using VGG-19 over the ImageNet dataset for images featuring multiple objects. While ScoreCAM exhibits inconsistent behavior, ScoreCAM++ successfully captures all the objects.}
 \label{fig:multiple_obj}

\end{figure}

\subsection{Results on ImageNet Dataset}

On the ImageNet dataset, for the ResNet-18 model, all methods generate well-structured class activation maps that cover most, if not all, of the relevant contexts and features. This observation is supported by Fig.~\ref{fig:resnet_imagenet}. However, the scenario shifts for VGG-19, where a diverse mapping pattern is exhibited across all the methods. The class activation maps produced by ScoreCAM++ stand out in contrast to other methods, as evident from Fig.~\ref{fig:vgg_imagenet}, and further reinforced by the data in Table~\ref{tab:vgg_imagenet}. In the case of the ResNet-18 architecture, ScoreCAM++ yields an average drop of 4.17\%, an average increase of 43.75\%, and a win percentage of 62.15\%, as evidenced in Table~\ref{tab:res_imagenet}. The results for VGG-19 are provided in Table~\ref{tab:vgg_imagenet}.

The mapping patterns of GradCAM and GradCAM++ are mostly identical for most images, with GradCAM++ sometimes deviating to provide a more focused explanation map. However, ScoreCAM exhibits some level of inconsistency, generating maps that are occasionally similar to GradCAM and GradCAM++, but, at times, diverging entirely to produce superior maps. Our proposed method consistently enhances GradCAM and GradCAM++ for most images, although the extent of improvement varies. In some cases, the improvements are so marginal that they appear nearly identical. The metrics reveal that both GradCAM and GradCAM++ generally exhibit slightly higher average drop values than their modified counterparts. However, the win percentage metric indicates a more notable increase, offering valuable insight into the effectiveness of these modifications in enhancing existing methods. Regarding metrics, we observe that GradCAM and GradCAM++ almost always have higher average drop values than their modified counterparts, even if only by a marginal amount.

\subsection{Results Analysis}
It is important to note that removing or adding a larger region may not necessarily improve performance metrics (i.e., Average Drop Percentage, Increase in Confidence, and Win Percentage), as evident in Figures~\ref{fig:dog_cat_vgg}, \ref{fig:resnet_imagenet}, and \ref{fig:vgg_imagenet}, where AugGradCAM++ produces larger activated regions compared to our approach (ScoreCAM++), but performs poorly in terms of evaluation metrics (see Tables~\ref{tab:res_dogcat}, \ref{tab:vgg_dogcat}, \ref{tab:res_imagenet}, \ref{tab:vgg_imagenet}). Through qualitative and quantitative results, we empirically demonstrate that our approach is capable of identifying all crucial regions in the image that explain the model's behaviour.

The characteristics of the Tanh activation function (with a range of $[-1,1]$) help in gating the non-important values towards 0 or -1 for lower or negative priority. Simultaneously, it elevates the higher priority regions towards 1. Our method establishes a clear distinction between high and low-priority values. Consequently, we obtain the resulting confidence score for the class based on the most crucial information from the activation layer. Therefore, we do not necessarily achieve a sparser heatmap. Furthermore, a sparser heatmap wouldn't necessarily be more faithful, as demonstrated by our experimental results (refer to Tables~\ref{tab:res_dogcat}, \ref{tab:vgg_dogcat}, \ref{tab:res_imagenet}, \ref{tab:vgg_imagenet} and Figures~\ref{fig:dog_cat_vgg}, \ref{fig:resnet_imagenet}, \ref{fig:vgg_imagenet}). Here, we observe that LayerCAM shows a highly sparse value but a highly degraded metric. Also, the highly non-sparse attention map AugGradCAM++ does not lead to an improvement in the metric. The proposed model captures the most relevant regions (neither sparse nor dense), which outperform all the metrics.

\section{Ablation study}
We conducted several ablation experiments to demonstrate the effectiveness of our proposed explainable approach.

\subsection{Comparison among Different Activation Functions}
In this section, we demonstrate that tanh yields the best quantitative and qualitative results compared to different types of activation functions. Table~\ref{tab:diff_act} provides a relative comparison among the five standard activation functions, clearly showing that \emph{tanh} outperforms all the competitors. Consequently, we adopt the \emph{tanh} activation function in our approach. Fig.~\ref{fig:diff_act} visually compares the explanation maps generated by ScoreCAM++ with different types of activation functions. Figs.~\ref{fig:diff_act}(a)-(e) showcase the maps generated by the proposed method when employed with various activation functions, namely, ReLU, Sigmoid, Swish, and Mish. These maps reveal discrepancies and a lack of clarity in precisely identifying the most salient region of interest within the image. Notably, the explanation maps generated using Swish and Mish exhibit a high degree of resemblance. On the other hand, Fig.~\ref{fig:diff_act}(f) illustrates the maps generated using the proposed \emph{tanh} activation function. It is evident from these figures that employing \emph{tanh} activation produces explanation maps characterized by visual coherence, alignment with the model's behavior, and an effective representation of key areas crucial for accurate recognition.

\subsection{Evaluation on Images with Multiple Objects}
As depicted in Fig.~\ref{fig:multiple_obj}, the visual explanations generated by ScoreCAM++ using VGG-19 over the ImageNet dataset for images containing multiple objects successfully capture all the objects. In contrast, ScoreCAM exhibits inconsistent behavior and may not consistently capture all the objects.

\subsection{Importance of Scaling Upsampled Layer}
We compare the performance of our proposed approach in two distinct cases: (1) when \emph{tanh} activation is applied only in place of normalization, and (2) when \emph{tanh} activation is used both in place of normalization and on the upsampled activation layer before multiplying it with the score. The relative comparison between these two cases is presented in Table~\ref{tab:tanh_ablation}. Table~\ref{tab:tanh_ablation} demonstrates that the most optimal results are obtained when \emph{tanh} is used for both normalization and scaling the upsampled activation layer before multiplying it with the score. This configuration showcases an average drop of 4.28\% and an average increase in confidence of 43.13\%. These results strongly support our idea that the combined utilization of \emph{tanh} for replacing normalization and scaling the activation layer significantly enhances the reliability and effectiveness of our approach.

\begin{table}[t]

\centering
\caption{Ablation experiments using VGG-19 over the ImageNet dataset involving two cases: (1) \emph{tanh} activation applied only in place of normalization, and (2) \emph{tanh} activation applied both in place of normalization and to the upsampled activation layer before multiplying with the score.}
\begin{tabular}{c|c|c}
\hline
& \begin{tabular}[c]{@{}c@{}}Normalization \\ + Upsampled\end{tabular} & Normalization \\ \hline
\begin{tabular}[c]{@{}c@{}}Average Drop \%\end{tabular}            & \textbf{4.28}                                                                    & 9.76          \\ \hline
\begin{tabular}[c]{@{}c@{}}Increase in Confidence\end{tabular} & \textbf{43.13}                                                                   & 28.07         \\ \hline
Win                                                                  & \textbf{79.86}                                                                   & 20.14         \\ \hline
\end{tabular}

\label{tab:tanh_ablation}

\end{table}

\begin{table}[t]
\caption{Average drop in Logit (higher value is better) corresponding to the correct class for various methods using the VGG-19 architecture over the ImageNet Dataset. AugGradCAM++ refers to Augmented GradCAM++.}
\label{tab:dropinlogit}
\addtolength{\tabcolsep}{-1.1mm}
\begin{tabular}{c|c|c|c}
\hline
GradCAM & GradCAM++ & XGradCAM & LayerCAM \\ \hline
7.81    & 8.04      & 7.92     & 7.99 \\ \hline
 ScoreCAM  &AugGradCAM++ & ScoreCAM++    \\ \hline
 7.89     & 9.08 & \textbf{9.65} \\ \hline
\end{tabular}
\end{table}

\subsection{Average Drop in Logit}
We conducted an ablation experiment where we removed the parts of the image indicated in the saliency map and then measured the drop in confidence (where the higher value is better) of the model corresponding to the correct class (using the model's logits as the benchmark). Averaged over 2000 randomly chosen images from the ImageNet Validation dataset, the drop in confidence (average drop in Logit corresponding to the correct class) in our method is the largest compared to other methods (see Table~\ref{tab:dropinlogit}). This demonstrates that the saliency map generated by our method accurately interprets the underlying model.

\section{Conclusion}
In this work, we introduce ScoreCAM++, a simple yet highly effective approach designed to generate smooth visual explanations for deep learning models through enhanced activation layer normalization and feature gating. By refining the normalization function and incorporating the \emph{tanh} activation, ScoreCAM++ effectively improves the interpretability of decision-making processes within complex deep learning models. Extensive experiments conducted on diverse datasets validate the superiority of ScoreCAM++ over existing methods. It consistently outperforms in interpreting CNN decisions, showcasing its effectiveness in providing clearer insights into model behaviors. By proposing simple yet impactful modifications to ScoreCAM, this work makes a valuable contribution to the field of Explainable AI. It offers a more trustworthy and effective means of visualizing CNN behaviors, facilitating a deeper understanding of their inner workings.


\appendix

\begin{table*}[t]
\centering
\caption{Results obtained using ViT architecture over ImageNet Dataset for various methods. Average Drop\%: lower is better. Increase in Confidence and Win\%: higher is better.}

\resizebox{\textwidth}{!}{
\begin{tabular}{|c|c|c|c|c|c|c|}
\hline
                       & GradCAM & GradCAM++ & XGradCAM & LayerCAM & ScoreCAM & ScoreCAM++     \\ \hline
Average Drop \%        & 48.99   & 56.90     & 45.63    & 18.17    & 19.31    & \textbf{6.52}  \\ \hline
Increase in Confidence & 9.85    & 7.30      & 7.75     & 18.50    & 29.75    & \textbf{36.35} \\ \hline
Win \%                 & 5.35    & 4.60      & 5.40     & 17.95    & 27.75    & \textbf{38.95} \\ \hline
\end{tabular}}
\label{tab:vit}

\end{table*}

\section{Evaluation Metrics}
\label{sec:metrics}

\subsection{Average Drop Percentage}
A good model captures the important features of an image. When we classify an image, the probability calculated for the correct class is expected to drop if we provide only a portion of the image compared to the whole image due to the loss of context. We use this phenomenon to check the performance of each model by comparing the confidence when the entire image is passed to the classifier vs. when passing only the explanation map. A better model will capture more relevant parts and thus preserve more relevant context, reducing the relative drop in confidence. We compute this metric as the average \% drop in the model’s confidence for a particular class in an image when having only the $\text{explanation map}$. The Average Drop Percentage is expressed as:

\begin{equation}
\text{Average Drop} = \left( \sum_{i=1}^{N} \frac{\max(0, y_{i}^c - O_{i}^c)}{y_{i}^c}\right) \times 100
\end{equation}
where $y_{i}^c$ represents the model’s output score (confidence) for class $c$ of the $i^{th}$ sample, and $O_{i}^c$ is the same model’s confidence in class $c$ for the $i^{th}$ sample with only the explanation map region as input. We use $\max$ in the numerator to handle cases where $O_{i}^c > y_{i}^c$. This value is computed per image and averaged over the entire dataset.

\subsection{Increase in Confidence}
Contrary to previous expectations, there must be scenarios where removing the context helps boost prediction confidence, especially when the context has noise or unnecessary distractions. Keeping in mind such cases, here we calculate the number of times passing only the explanation map to the classifier has boosted its performance compared to passing the entire image.

Formally, the increase in confidence is defined as:
\begin{equation}
\text{Increase in Confidence} = \frac{1}{N} \sum_{i=1}^{N} 1_{\left(y_{i}^c < O_{i}^c\right)} \times 100,
\end{equation}
where $1_{x}$ is an indicator function that returns 1 when the argument is true. All other notations are the same as defined for the previous metric.

\subsection{Win Percentage}
In addition to the average increase and average drop, we calculated the win metric. The win metric for a particular model calculates the number of times the drop in confidence by that model was the least among the other competing models. Then, the values are expressed as a percentage.

\section{Experimental Results using Vision Transformer (ViT) on ImageNet Dataset}
To further assess the superiority of the proposed method, we conducted experiments using the Vision Transformer (ViT) architecture. Our proposed architecture, ScoreCAM++, outperforms all other architectures while explaining the Vision Transformer in all three given metrics: Average Drop, Increase in Confidence, and Win percentages (see Table~\ref{tab:vit}). It also surpasses the second-best method in our experiment, ScoreCAM, by a significant margin.

\section{Implementation Details}
\label{sec:implem}
In the conducted experiments involving the ImageNet dataset, we utilized pretrained VGG-19 and ResNet-18 models obtained from the PyTorch model zoo as our base models. Input images were resized to dimensions of $224 \times 224 \times 3$, normalized to the range [0, 1], and underwent additional processing involving mean vector normalization $[0.485, 0.456, 0.406]$ and standard deviation vector normalization $[0.229, 0.224, 0.225]$. No further preprocessing steps were applied to the images.

For experiments related to the Cats and Dogs dataset, the same pretrained VGG-19 and ResNet-18 models from the PyTorch model zoo were employed as base models. These models underwent additional training on the Cats and Dogs dataset for 100 epochs, utilizing CrossEntropy as the loss function. The image pre-processing remained consistent with that employed for the ImageNet Dataset.

For experiments related to the Vision Transformer, we used the pretrained \texttt{vit\_b16} model from the PyTorch model zoo. The ImageNet dataset was used for this experiment. The image pre-processing remained consistent with the previous experiments.

In terms of performance evaluation, the models were trained with batch sizes of 64. The Adam optimizer with a learning rate of 0.0001 was employed, and the models were trained for 100 epochs. The implementation of the proposed method was carried out using PyTorch 2.0.1 and Python 3.11.3. The hardware configuration comprised an AMD EPYC 7543 and an NVIDIA RTX A5000 with 256GB RAM.

\bibliographystyle{elsarticle-num}

\bibliography{cas-refs}

\end{document}